%% file: chen2021ral-iros.tex
\documentclass[letterpaper, 10 pt, journal, twoside]{ieeetran}

\input{stachnisslab-latex}

\input{stachnisslab-math}

\usepackage{subfigure}
\usepackage{pifont}
\usepackage{amssymb}
\usepackage{multirow}
\usepackage{graphicx}
\graphicspath{{pics/}}

\title{\LARGE \bf Moving Object Segmentation in 3D LiDAR Data:\\ A Learning-based Approach Exploiting Sequential Data}

\author{Xieyuanli Chen \and Shijie Li \and Benedikt Mersch \and Louis Wiesmann \and J\"urgen Gall \and Jens Behley \and Cyrill Stachniss
	\thanks{Manuscript received: February 24, 2021; Revised May 19, 2021; Accepted June 17, 2021.
	This paper was recommended for publication by Editor Javier Civera upon evaluation of the Associate Editor and Reviewers' comments.}
	\thanks{This work has partially been funded by the Deutsche Forschungsgemeinschaft (DFG, German Research Foundation) under Germany's Excellence Strategy, EXC-2070 -- 390732324 -- PhenoRob, under GA 1927/5-2 (FOR 2535), by the EC within Horizon Europe, grant agreement no.~101017008 (Harmony), and by the Chinese Scholarship Committee.
	All authors are with the University of Bonn, Germany. {\tt\footnotesize xieyuanli.chen@igg.uni-bonn.de}}%
\thanks{Digital Object Identifier (DOI): 10.1109/LRA.2021.3093567.}
}

\begin{document}
\maketitle

\markboth{IEEE Robotics and Automation Letters. Preprint Version. June, 2021}
{Chen \MakeLowercase{\textit{et al.}}: Moving Object Segmentation in 3D LiDAR Data}

\begin{abstract}
	The ability to detect and segment moving objects in a scene is essential for building consistent maps, making future state predictions, avoiding collisions, and planning.
	In this paper, we address the problem of moving object segmentation from 3D LiDAR scans. We propose a novel approach that pushes the current state of the art in LiDAR-only moving object segmentation forward to provide relevant information for autonomous robots and other vehicles. Instead of segmenting the point cloud semantically, \ie, predicting the semantic classes such as vehicles, pedestrians,  roads, etc., our approach accurately segments the scene into moving and static objects, \ie, also distinguishing between moving cars vs.~parked cars. Our proposed approach exploits sequential range images from a rotating 3D LiDAR sensor as an intermediate representation combined with a convolutional neural network and runs faster than the frame rate of the sensor. We compare our approach to several other state-of-the-art methods showing superior segmentation quality in urban environments. Additionally, we created a new benchmark for LiDAR-based moving object segmentation based on SemanticKITTI. We published it to allow other researchers to compare their approaches transparently and we furthermore published our code.
\end{abstract}

\section{Introduction}
\label{sec:intro}

\IEEEPARstart{T}{he} ability to identify which parts of the environment are static and which ones are moving is key to safe and reliable autonomous navigation. It supports the task of predicting the future state of the surroundings, collision avoidance, and planning. This knowledge can also improve and robustify pose estimation, sensor data registration, and simultaneous localization and mapping~(SLAM). Thus, accurate and reliable moving object segmentation~(MOS) in sensor data at frame rate is a crucial capability supporting most autonomous mobile systems. Depending on the application domain and chosen sensor setup, moving object segmentation can be a challenging task.

\begin{figure}[t]
	\centering
	\fontsize{8}{8}\selectfont
	\def\svgwidth{\linewidth}
	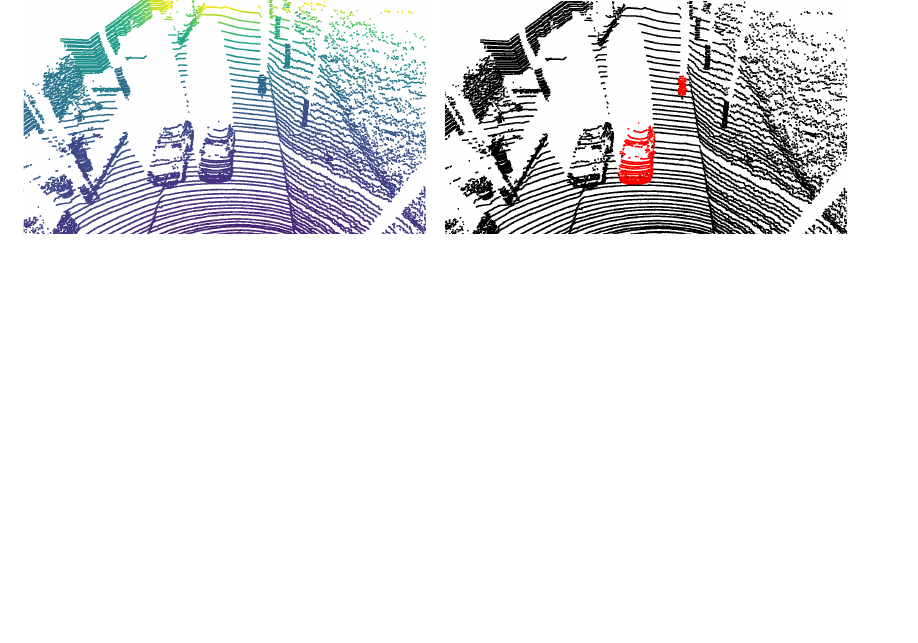
	\caption{Moving object segmentation using our approach. Our method can detect and segment the currently moving objects given sequential point cloud data exploiting its range projection. Instead of detecting all \emph{potentially movable} objects such as vehicles or humans, our approach distinguishes between \emph{actually moving} objects (labeled in red) and static or non-moving objects (black) in the upper row. At the bottom, we show the range image and our predictions in comparison to the ground truth labels.}
	\label{fig:motivation}
	\vspace{-0.3cm}
\end{figure}

In this work, we address the problem of moving object segmentation in 3D LiDAR data at sensor frame rate in urban environments. Instead of detecting all \emph{theoretically movable} objects such as vehicles or humans, we aim at separating the \emph{actually moving} objects such as driving cars from static or non-moving objects such as buildings, parked cars, etc. See \figref{fig:motivation} for an example scene and our segmentation. Moving objects are colored in red. We propose a novel approach based on convolutional neural networks~(CNNs) to explicitly address the MOS problem for 3D LiDAR scans. We exploit range images as an intermediate representation, which is a natural representation of the scan from a rotating 3D LiDAR such as a Velodyne or Ouster sensors. Based on this comparably light-weight representation, we can directly exploit the existing range-image-based semantic segmentation networks as proposed by Milioto \etal~\cite{milioto2019iros}, Cortinhal \etal~\cite{cortinhal2020iv}, and Li \etal~\cite{li2020multi} to tackle the MOS problem. Most of such existing LiDAR-based semantic segmentation networks predict the semantic labels of a point cloud, \eg vehicle, building, road, etc. They do not distinguish between actually moving objects, like moving cars, and static objects, like parked cars and also buildings, etc. We are making this distinction and are exploiting sequences of range images, allowing for an effective moving object segmentation targeted to autonomous vehicles. Our main application focus is perception for self-driving cars in outdoor scenes, but the method itself is not restricted to this domain.

This paper's main contribution is a novel method based on CNNs using range images generated from 3D LiDAR scans together with the residual images generated from past scans as inputs and outputs for each range measurement in the current frame a label indicating if it belongs to a moving object or not. By combining range images and residual images our network exploits the temporal information and can differentiate between moving and static objects as shown in~\figref{fig:motivation}. 
For training, we reorganize the SemanticKITTI~\cite{behley2019iccv} dataset and merge the original labels into two classes, moving and static, by exploiting the existing annotations of moving traffic participants. Furthermore, our approach runs faster than the sensor frame rate, \ie, 10\,Hz for a typical rotating 3D LiDAR sensor. Comparisons with multiple existing methods suggest that the proposed approach leads to more accurate moving object segmentation. In sum, we make two key claims: First, our approach is able to achieve moving object segmentation using only 3D LiDAR scans and runs faster than the sensor frame rate of 10\,Hz. Second, it improves the moving object segmentation performance by incorporating residual images in addition to the current scan and outperforms several state-of-the-art networks. To allow for as easy as possible comparisons and support future research, we propose and release a moving object segmentation benchmark\,\footnote{\label{myfootnote}http://bit.ly/mos-benchmark}, including a hidden test set, based on the SemanticKITTI dataset and we release the source code of our approach\,\footnote{\label{footnode:code}https://github.com/PRBonn/LiDAR-MOS}. We also provide a short video\footnote{\label{footnode:code}https://youtu.be/NHvsYhk4dhw} illustrating the capabilitiy of our method.

\section{Related Work}
\label{sec:related}

While there has been a large interest in vision-based~\cite{mcmanus2013icra, barnes2018icra, patil2020cvpr}, radar-based~\cite{aldera2019icra} and vision and LiDAR combined~\cite{yan2014threedv, postica2016iros} moving object segmentation approaches, we concentrate here on approaches using only LiDAR sensors.
Below, we distinguish between map-based and map-free approaches. 

\textbf{Map-based approaches.}
Most of the existing LiDAR-based approaches target the cleaning of a point cloud map. These methods mostly run offline and rely on a prebuilt map.
Some methods use time-consuming voxel ray casting and require accurately aligned poses to clean the dense terrestrial laser scans~\cite{gehrung2017isprsannals, schauer2018ral}.
To alleviate the computational burden, visibility-based methods have been proposed~\cite{pomerleau2014icra, xiao2015jprs}. These types of methods associate a query point cloud to a map point within a narrow field of view, \eg cone-shaped used by Pomerleau \etal~\cite{pomerleau2014icra}.
Recently, Pagad \etal~\cite{pagad2020icra} propose an occupancy map-based method to remove dynamic points in LiDAR scans. They first build occupancy maps using object detection and then use the voxel traversal method to remove the moving objects.
Kim \etal~\cite{kim2020iros} propose a range-image-based method, which exploits the consistency check between the query scan and the pre-built map to remove dynamic points and uses a multi-resolution false prediction reverting algorithm to refine the map.
Even though such map-based methods can separate moving objects from the background, they need a pre-built and cleaned map and therefore usually can not achieve online operation.

\begin{figure*}[t]
	\centering
	\fontsize{8}{8}\selectfont
	\def\svgwidth{0.95\linewidth}
	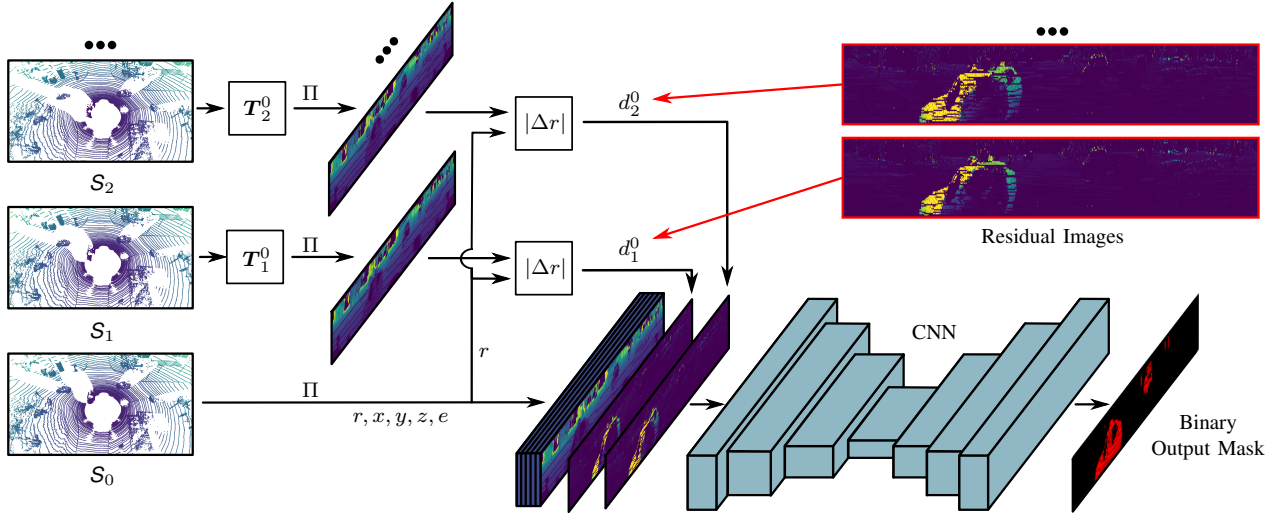
	\caption{Overview of our method. We use the range projection-based representation of LiDAR scans to achieve online moving object segmentation. Given the current scan $\m{S}_0$, we generate residual images from previous scans $\{\m{S}_{i}\}_{i=1}^N$ to explore the sequential information. This is by transforming them to the current viewpoint with a homogeneous transformation matrix $\v{T}_i^0$ estimated from a SLAM or sensor-based odometry, projecting them to the range representation with a mapping $\Pi$ and subtracting them from the current scan's range image. The residual images are then concatenated with the current range image and used as input to a fully convolutional neural network. Trained with the binary labels, the proposed method can separate moving and static objects.}
	\label{fig:overview}
\end{figure*}

\textbf{Map-free approaches.}
Recently, LiDAR-based semantic segmentation methods operating only on the sensor data have achieved great success~\cite{milioto2019iros, cortinhal2020iv, li2020multi}.
Wang \etal~\cite{wang2012icra} tackle the problem of segmenting things that could move from 3D laser scans of urban scenes, \eg cars, pedestrians, and bicyclists.
Ruchti and Burgard \etal~\cite{ruchti2018icra} also propose a learning-based method to predict the probabilities of potentially movable objects.
Dewan \etal~\cite{dewan2016iros} propose a LiDAR-based scene flow method that estimates motion vectors for rigid bodies.
Based on that, they developed both semantic classification and segmentation methods~\cite{dewan2017iros, dewan2020icra}, which exploit the temporally consistent information from the sequential LiDAR scans.
Bogoslavskyi and Stachniss~\cite{bogoslavskyi2017pfg} propose a class-agnostic segmentation method for 3D LiDAR scans that exploits range images to enable online operation and leads to more coherent segments, but does not distinguish between moving and non-moving objects.

Semantic segmentation can be seen as a relevant step towards moving object segmentation. Most existing semantic segmentation methods, however, only find \textit{movable} objects, \eg vehicles and humans, but do not distinguish between actually moving objects, like driving cars or walking pedestrians, and non-moving/static objects, like parked cars or building structures. The most similar work to ours is the one by Yoon \etal~\cite{yoon2019cvr}, which also detects dynamic objects in LiDAR scans without using a pre-built map. It exploits heuristics, \eg the residual between LiDAR scans, free space checking, and region growing to find moving objects.
There are also multiple 3D point cloud-based semantic segmentation approaches~\cite{thomas2019iccv,tang2020eccv,shi2020cvpr}, which also perform well in semantic segmentation tasks.
Among them, Shi \etal~\cite{shi2020cvpr} exploit sequential point clouds and predict moving objects.
However, based on networks operating directly on point clouds, these methods are usually heavy and difficult to train.
Furthermore, most of them are both time-consuming and resource-intensive, which might not be applicable for autonomous driving.

Our method is also based on neural networks and we investigate the usage of three recent range projection-based semantic segmentation methods proposed by Milioto \etal~\cite{milioto2019iros}, Cortinhal \etal~\cite{cortinhal2020iv}, and Li \etal~\cite{li2020multi} to tackle MOS with the prospect of real-time capability and operation beyond the frame rate of the LiDAR sensor.
Our method does not rely on a pre-built map and operates online, \ie, uses only LiDAR scans from the past.
We exploit residuals between the current frame and the previous frames as an additional input to the investigated semantic segmentation networks to enable class-agnostic moving object segmentation.
Note that the proposed architecture does not depend on a specific range projection-based semantic segmentation architecture. By training the network with proposed new binary masks, our method distinguishes between moving cars and parked cars in an end-to-end fashion.

\section{Our Approach}
\label{sec:main}

The goal of our approach is to achieve accurate and fast moving object segmentation~(MOS) for LiDAR scans to enable autonomous mobile systems to make decisions in a timely manner.
\figref{fig:overview} shows a conceptual overview of our proposed method.
To achieve online MOS, we first project the point clouds into range image representation (see~\secref{sec:range}).
To separate moving and non-moving objects, we then exploit sequential information (see~\secref{sec:sequential}) computing residuals between the current and the previous scans (see~\secref{sec:residual}). We finally concatenate them together with the range information as the input for a segmentation CNN (see~\secref{sec:cnn}).
In addition, we propose a novel MOS benchmark based on SemanticKITTI (see~\secref{sec:benchmark}) to train and evaluate MOS methods.

\subsection{Range Image Representation}
\label{sec:range}

In line with prior work~\cite{milioto2019iros, li2020multi, chen2021icra}, we use a range projection of a point cloud to generate an image representation.
Specifically, we convert each LiDAR point $\v{p}=(x, y, z)$ via a mapping $\Pi: \RR^3 \mapsto \RR^2$ to spherical coordinates, and finally to image coordinates, as defined by
\begin{align}
	\left( \begin{array}{c} u \vspace{0.0em} \\ v \end{array}\right) & = \left(\begin{array}{cc} \frac{1}{2}\left[1-\arctan(y, x) \, \pi^{-1}\right]~\,~w   \vspace{0.5em} \\
			\left[1 - \left(\arcsin(z\, r^{-1}) + \mathrm{f}_{\mathrm{up}}\right) \mathrm{f}^{-1}\right] \, h\end{array} \right), \label{eq:projection}
\end{align}
where $(u,v)$ are image coordinates, $(h, w)$ are the height and width of the desired range image representation, $\mathrm{f}~{=}~\mathrm{f}_{\mathrm{up}}~{+}~\mathrm{f}_{\mathrm{down}}$ is the vertical field-of-view of the sensor, and $r~{=}~||\v{p}_i||_2$ is the range of each point.
This procedure results in a list of~$(u,v)$ tuples containing a pair of image coordinates for each $\v{p}_i$. Using these indices, we extract for each $\v{p}_i$, its range~$r$, its $x$, $y$, and~$z$ coordinates, and its remission~$e$, and store them in the image.
Thus, each pixel can store more than only a range.
Consequently, we can easily exploit extra information and add this as extra channels.
Therefore, we can directly feed this information to existing networks without changing the architectures, which makes our method easily transferred to other new architectures.
To show this capability, we test our method with three different segmentation networks in this paper.

\subsection{Sequence Information}
\label{sec:sequential}

We aim at segmenting moving objects online, \ie, only using the current and recent LiDAR scans, such that one can exploit the information for odometry estimation in a SLAM pipeline and potentially remove dynamics from a map representation.
We assume that we are given a time series of~$N$ LiDAR scans in the SLAM history, denoted by \mbox{$\m{S}_{j}~=\{\v{p}_{i} \in \mathbb{R}^4\} $} with~$M$ points represented as homogeneous coordinates, \ie, $\v{p}_i = (x,y,z,1)$.
We denote the current LiDAR scan by $\m{S}_0$ and the sequence of $N$ previously scans by $\m{S}_{j}$ with $1<j<N$.
The \emph{estimated} $N$ consecutive relative transformations from the SLAM\,/\,odometry approach, $\v{T}^{N-1}_{N}, \dots, \v{T}^{0}_{1}$, between the $N+1$ scanning poses, represented as transformation matrices, \ie, $\v{T}^l_k \in \mathbb{R}^{4\times4}$, are also assumed to be available. Given the  estimated relative poses between consecutive scans, we can transform points from one viewpoint to another.
We denote the $k^\text{th}$ scan transformed into the $l^\text{th}$ scan's coordinate frame by
\begin{align}
	\m{S}^{k \rightarrow l} & =\{\v{T}^l_k\v{p}_{i} | \v{p}_i \in \m{S}_k\}, 
	 \label{eq:transformed_scan}
\end{align}
with $\v{T}^l_k = \prod^{l+1}_{j = k} \v{T}^{j-1}_j$.

\subsection{Residual Images}
\label{sec:residual}

Inspired by Wang \etal~\cite{wang2018tpami}, who exploit the difference between RGB video frames for action recognition,
we propose to use LiDAR-based residual images together with pixel-wise binary labels on the range image to segment moving objects. Combining the current sensor reading and residual images, we can employ existing segmentation networks to distinguish between pixels on moving objects and background by leveraging the temporal information inside the residual images.

\begin{figure}[t]
	\centering
	\def\svgwidth{0.95\linewidth}
	\fontsize{8}{8}\selectfont
	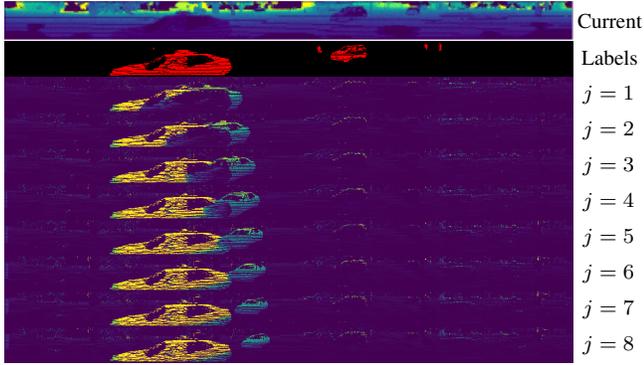
	\caption{Residual images, where $j$ means the residual image generated between the current frame and the last $j$-th frame. We can see the continuous discrepancy in the residual images due to the motion of the moving car.}
	\label{fig:diff}
\end{figure}

To generate the residual images and later fuse them into the current range image, transformation, and re-projection are required. To realize this, we propose a three-step procedure:
First, we compensate for the ego-motion by transforming the previous scans into the current local coordinate system given the transformation estimates as defined in \eqref{eq:transformed_scan}. 
Next, the transformed past scans~$\m{S}^{k \rightarrow l}$ are re-projected into the current range image view using ~\eqref{eq:projection}.
We compute the residual $d^{l}_{k,i}$ for each pixel $i$ by computing the normalized absolute difference between the ranges of the current frame and the transformed frame by
\begin{align}
	d^{l}_{k,i} =  \frac{\abs{r_i - r^{k \rightarrow l}_i}}{r_i},
\end{align}
where $r_i$ is the range value of $\v{p}_{i}$ from the current frame located at image coordinates $(u_i,v_i)$ and $r^{k \rightarrow l}_i$ is the corresponding range value from the transformed scan located at the same image pixel.
We only calculate the residual for the valid pixels that contain measurements and set the residual to zero for the invalid pixels.
Examples of such residual images are depicted in~\figref{fig:diff}.
We can see that due to the motion of objects in the scene, \eg the moving car, the displacement between these points in the common viewpoint is relatively large compared to the static background.
However, there are ambiguities, since the large residual patterns appear twice for one moving object, while for the slowly moving objects the residual patterns are not obvious. Therefore, directly using residual images for moving object segmentation does not lead to a great performance. It, however, provides a valuable cue for moving objects and can guide the network to separate moving and non-moving objects.

In the end, the residual images are concatenated with the current range image as extra channels where range image provides spatial information and residual image encodes temporal information.
Each pixel $(u_i,v_i)$ in the fused range image then contains a vector of different types of information $(x_i, y_i, z_i, r_i, e_i, d_{1,i}^0, ..., d_{j,i}^0, ...,d_{N-1,i}^0)$, where $d_{j}^0$ is the residual image calculated between the last $j^\text{th}$ frame and the current frame.

\subsection{Range Projection-based Segmentation CNNs}
\label{sec:cnn}

In this paper, we do not design a new network architecture but reuse networks that have been successfully applied to LiDAR-based semantic segmentation in the past. We adopt and evaluate three popular networks, namely SalsaNext~\cite{cortinhal2020iv}, RangeNet++~\cite{milioto2019iros}, and MINet~\cite{li2020multi}, for MOS. SalsaNext and RangeNet++ are encoder-decoder architectures with a solid performance and MINet uses a light-weight and efficient multi-path architecture. After the segmentation, a fast GPU-based k-Nearest-Neighbor search over the point cloud is used to remove artifacts produced by the range projection~\cite{milioto2019iros}. 
All methods are state-of-the-art range projection-based LiDAR semantic segmentation networks, comparably light-weight, and can achieve real-time operation, \ie, run faster than the frame rate of the employed LiDAR sensor, which is $10$\,Hz for common Ouster and Velodyne scanners. For more detailed information about each network, we refer to the original papers~\cite{milioto2019iros, cortinhal2020iv, li2020multi}.

Instead of changing the architecture of these segmentation networks, we directly feed them with the fused range images plus the residual information, retrain the network and evaluate their performance with our MOS benchmark proposed in~\secref{sec:benchmark}. Using our proposed residual image approach, all segmentation networks show a large improvement in moving object segmentation as shown in~\secref{sec:ablation}.
For training, we use the same loss functions as used in the original segmentation methods, while mapping all classes into two per-point classes, moving and non-moving.

\subsection{Moving Object Segmentation Benchmark}
\label{sec:benchmark}
Large datasets for LiDAR-based odometry, object detection, and tracking, like the KITTI Vision Benchmark~\cite{geiger2012cvpr}, and semantic segmentation, panoptic segmentation, and scene completion like SemanticKITTI~\cite{behley2019iccv} are available and widely used. There are, however, not many datasets and benchmarks available for 3D LiDAR-based moving object segmentation. With this work, we also aim at covering this gap with a novel benchmark task for MOS.

Our proposed MOS benchmark is based on SemanticKITTI. It uses the same split for training and test set as used in the original odometry dataset, where sequences 00 to 10 are used for training and sequences 11 to 21 are used as a test set.
SemanticKITTI contains in total 28 semantic classes such as vehicles, pedestrians, buildings, roads, etc. and distinguishes between moving and non-moving vehicles and humans.
In the proposed MOS benchmark, we manually reorganize all the classes into only two types: moving and non-moving/static objects. The actually moving vehicles and humans belong to moving objects and all other classes belong to the non-moving/static objects.

For quantifying the MOS performance, we use the commonly applied Jaccard Index or intersection-over-union (IoU) metric~\cite{everingham2010ijcv} over moving objects, which is given by
\begin{align}
	\text{IoU} & = \frac{\text{TP}}{\text{TP} + \text{FP} + \text{FN}}, \label{eq:miou}
\end{align}
where $\text{TP}$, $\text{FP}$, and $\text{FN}$ correspond to the number of true positive, false positive, and false negative predictions for the moving class.

\section{Experimental Evaluation}
\label{sec:exp}

This paper focuses on moving object segmentation from 3D LiDAR scan sequences. We present our experiments to show the capabilities of our method and to support our key claims, that our approach:
(i) achieves moving object segmentation using only 3D LiDAR scans and runs faster than the sensor frame rate of 10\,Hz and
(ii) improves the moving object segmentation performance by using residual images, and outperforms several state-of-the-art networks.

We evaluate all the methods on our proposed MOS benchmark.
We use the odometry information provided by SemanticKITTI, which are estimated with a LiDAR-based SLAM system, SuMa~\cite{behley2018rss}. Aiming at an easy-to-integrate algorithm, we stick to the original setup while only changing the input and the output of the classification head into the proposed binary labels. We train each network using their specific training hyperparameters over 150 epochs on sequences 00-07 and 09-10 and keep sequence 08 as the validation set.
For more details on the training regime for each network, we refer to the original papers~\cite{milioto2019iros, cortinhal2020iv, li2020multi}.

\subsection{Ablation Study on Input and Architecture}
\label{sec:ablation}
The first ablation study presented in this section is designed to support our claim that our approach is able to achieve moving object segmentation using only 3D LiDAR scans.
All the experiments in this section are evaluated on the validation set, \ie, sequence 08.

We test three different setups with three different networks, RangeNet++, SalsaNext, and MINet, for moving object segmentation as shown in~\tabref{tab:variants}.
The first setup is to train the three range projection-based networks directly with the labels for moving and non-moving classes.
The second setup is to attach the previous frames to the current frame as the input of the network resulting in $2\times 5$ input channels, as each image contains the coordinates~$(x,y,z)$, the range~$r$, and the remission~$e$ for each pixel.
The third setup is to concatenate the proposed residual images to the current frame as the input of the network and therefore the  input size is $5{+}N$, as detailed in \secref{sec:residual}.

As can be seen in~\tabref{tab:variants}, RangeNet++ and SalsaNext show a basic performance while MINet fails when training the network together with the binary labels and no additional inputs. Overall, the performance has space for improvements. This is probably due to the fact, that from one frame, the semantic segmentation networks cannot distinguish well between the moving and static objects from the same general class, but may learn some heuristics, \eg that cars on the road are usually moving while those on parking lots are static, which can also be seen in the qualitative results~\figref{fig:qualitative}. A reason why MINet fails may be due to the lightweight architecture that is not capable of learning such heuristics.

In the second setup, we  directly combine two frames. Here, the networks can already gain some improvements in MOS, since they can obtain the temporal information from two scans. In this setting, MINet is also capable of predicting moving objects. In the third setup, the best MOS performance is achieved. We hypothesize that it is advantageous to give direct access to the residual information instead of the full range views. Given that most of the points are redundant in two successive frames and the input is large due to the concatenation, the networks need less time to extract the temporal information if the residuals are provided directly.
While a large enough network should be able to learn concepts as the difference between frames given enough time, it is generally advantageous to directly provide this information as also shown by Milioto \etal~\cite{milioto2019iros}.

\begin{figure}[t]
	\centering
	\includegraphics[width=0.48\linewidth]{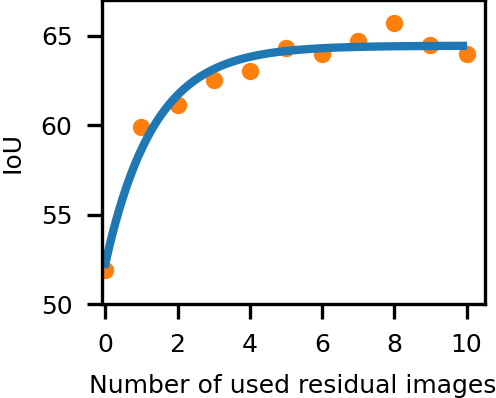}
	\includegraphics[width=0.48\linewidth]{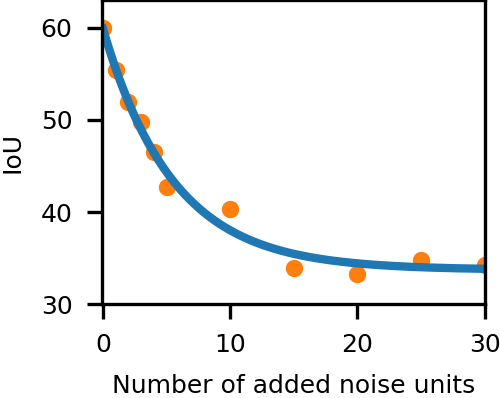}
	\caption{Ablation studies. The left figure shows the ablation study on the MOS performance \vs the number of residual images $N$. 
	The right figure shows the ablation study on the MOS performance \vs the number of added noise units to the poses during the inferring.
}
	\label{fig:residual}
\end{figure}

As shown in~\figref{fig:residual}, we provide two further ablation studies using SalsaNext as the segmentation network.
The left figure shows an ablation study on the number of residual images used for MOS Both ablation studies use SalsaNext as the segmentation network. We can see that $N=1$ residual image attains the biggest improvement in terms of MOS performance, while adding more residual images improves the MOS performance further with diminishing returns for $N>8$ residual images.
The figure on the right shows an ablation study on the MOS performance \vs the amount of noise added to the relative odometry poses used to generate the residual images.
We manually add noise to the the poses estimated by SLAM in $(x, y, \mathit{yaw})$
with a unit of $(0.1\,m, 0.1\,m, 1^\circ)$ to see how the pose estimations influence our method during inferring. As can be seen, the MOS performance will drop due to the noisy poses. However, when the added noises are larger than $20$~units, $(2\,m, 2\,m, 20^\circ)$, the network may ignore the noisy residual images and the MOS performance will not become worse. 

\begin{table}[t]
	\caption{Evaluating our method with three different networks}
	\centering

	\begin{tabular}{lccc}
		\toprule
		Input                 & RangeNet++ & MINet & SalsaNext \\
		\midrule
		One frame             & 38.9       & 9.1   & 51.9      \\
		Two frames            & 40.6       & 35.0  & 56.0      \\
		Residual frames (N=1) & 40.9       & 38.9  & 59.9      \\
		\bottomrule
	\end{tabular}

	\label{tab:variants}
\end{table}

\subsection{MOS Performance and Comparisons}

\begin{table}[t]
	\caption{MOS performance compared to the state of the art.}
	\centering

	\begin{tabular}{L{6cm}r}
		\toprule
		                                   & IoU  \\
		\midrule
		SalsaNext (moveable classes)       & 4.4  \\
		SalsaNext (retrained)              & 46.6 \\
		\midrule
		Residual                           & 1.9  \\
		Residual + RG                      & 14.1 \\
		Residual + RG + Semantics          & 20.6 \\
		\midrule
		SceneFlow                          & 4.8  \\
		SceneFlow + Semantics              & 28.7 \\
		\midrule
		SqSequence                         & 43.2 \\
		KPConv                             & 60.9 \\
		\midrule
		Ours (based on SalsaNext/N = 1)             & 52.0 \\
		Ours (based on SalsaNext/N = 8 + Semantics) & $\mathbf{62.5}$ \\
		\bottomrule
	\end{tabular}
	\vspace{-0.3cm}
	\label{tab:baselines}
\end{table}

The experiment presented in this section investigates the MOS performance of our approach. It supports the claim that our approach improves the MOS performance by using residual images and outperforms several state-of-the-art networks.
Since there are not many existing implementations for LiDAR-based MOS available, we choose several methods that have been used in similar tasks, \eg, semantic segmentation and scene flow, and modify them to achieve LiDAR-based MOS.
All the methods are evaluated on the test data of the proposed benchmark, \ie, sequences 11-21.

We analyze multiple alternative approaches. 
We start using an existing semantic segmentation network, \eg SalsaNext~\cite{cortinhal2020iv}, directly and label all the movable objects, \eg vehicles and humans, as moving objects while labeling other objects as static.
We name this method as \emph{SalsaNext (movable classes)}.
Here, we also show the results generated by the retrained SalsaNext with the proposed binary labels, named as \emph{SalsaNext (retrained)}.
Since the residual images can already point out rough positions of moving objects, here we also take it as a simple heuristic-based baseline, named as \emph{Residual}.
Inspired by Yoon \etal~\cite{yoon2019cvr}, we also re-implement the pure geometric heuristic-based method using residual information together with free space checking and region growing, named as \emph{Residual+RG}.

We furthermore compare our method also to the state-of-the-art scene flow method, FlowNet3D~\cite{liu2019cvpr}, referred to as \emph{SceneFlow}, which is a network estimating the translational flow vector for every LiDAR point given two consecutive scans as input.
We set a threshold on the estimated translation of each point to decide the label for each point, \ie, points with translations larger than the threshold are labeled as moving.
We fix the threshold based on the best MOS performance on the validation set.
We also compare our method to the state-of-the-art multiple point cloud-based semantic segmentation methods~\cite{thomas2019iccv,shi2020cvpr}, since they can also distinguish between moving and non-moving classes.

For the non-semantic-based methods, we additionally add semantic information by checking if the predicted moving objects are movable or not, and only label a point as moving if it is both predicted as moving by the original method and at the same time assigned to a movable object, \eg vehicles and humans.
The semantic information is generated using SalsaNext with the pre-trained weights provided by the original paper.
We identify the semantic-enhanced methods by adding \emph{+Semantics}.

We compare two setups of our method to all the above-mentioned methods.
For our methods, we choose SalsaNext as the base network as it shows the best performance in our ablation study.
In the first setup, we use only one residual image, $N=1$, to obtain the temporal information, and in the other setup, we use our best setup fixed on the validation sequence with $N=8$ residual images and semantic information to see the best performance of our method.

As shown in~\tabref{tab:baselines}, our residual image-based method with $N=1$ already outperforms most baselines, while being worse than KPConv, which is a dense multiple point clouds-based semantic segmentation method.
Due to heavy computation burden, it cannot achieve real-time performance.
When our method uses multiple residual images ($N=8$) together with semantic information, our method outperforms all other methods.

\begin{figure*}[t]
	\centering
	\def\svgwidth{0.90\linewidth}
	\fontsize{8}{8}\selectfont
	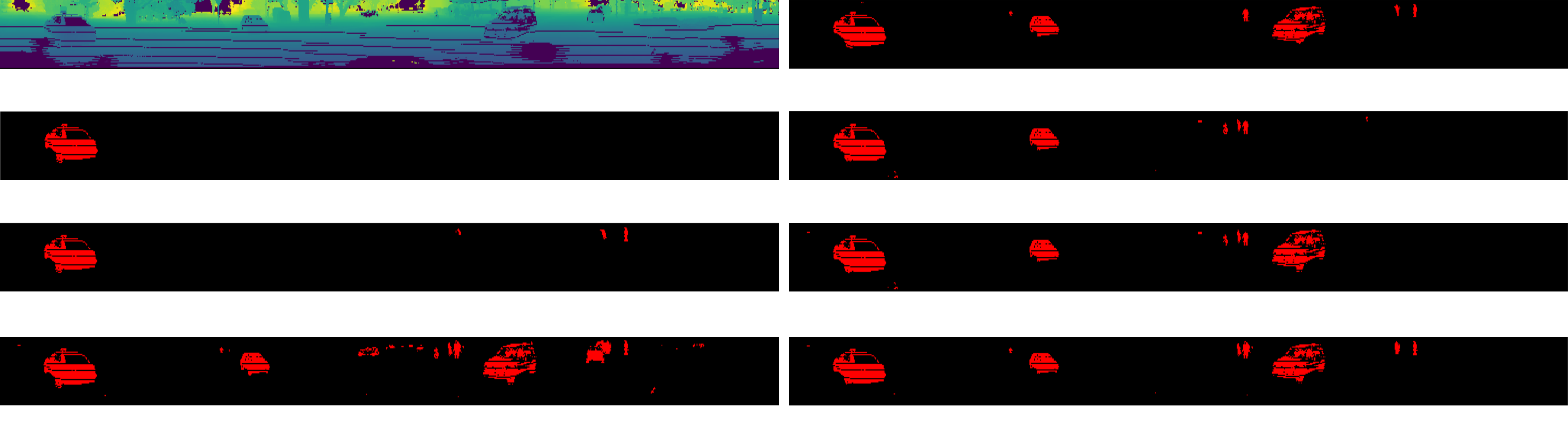
	\caption{Qualitative results with range projections, where red pixels correspond to moving objects.}
	\label{fig:qualitative}
\end{figure*}

\begin{figure*}[t]
	\centering
	\subfigure[Raw Point Cloud]{\includegraphics[width=0.42\linewidth]{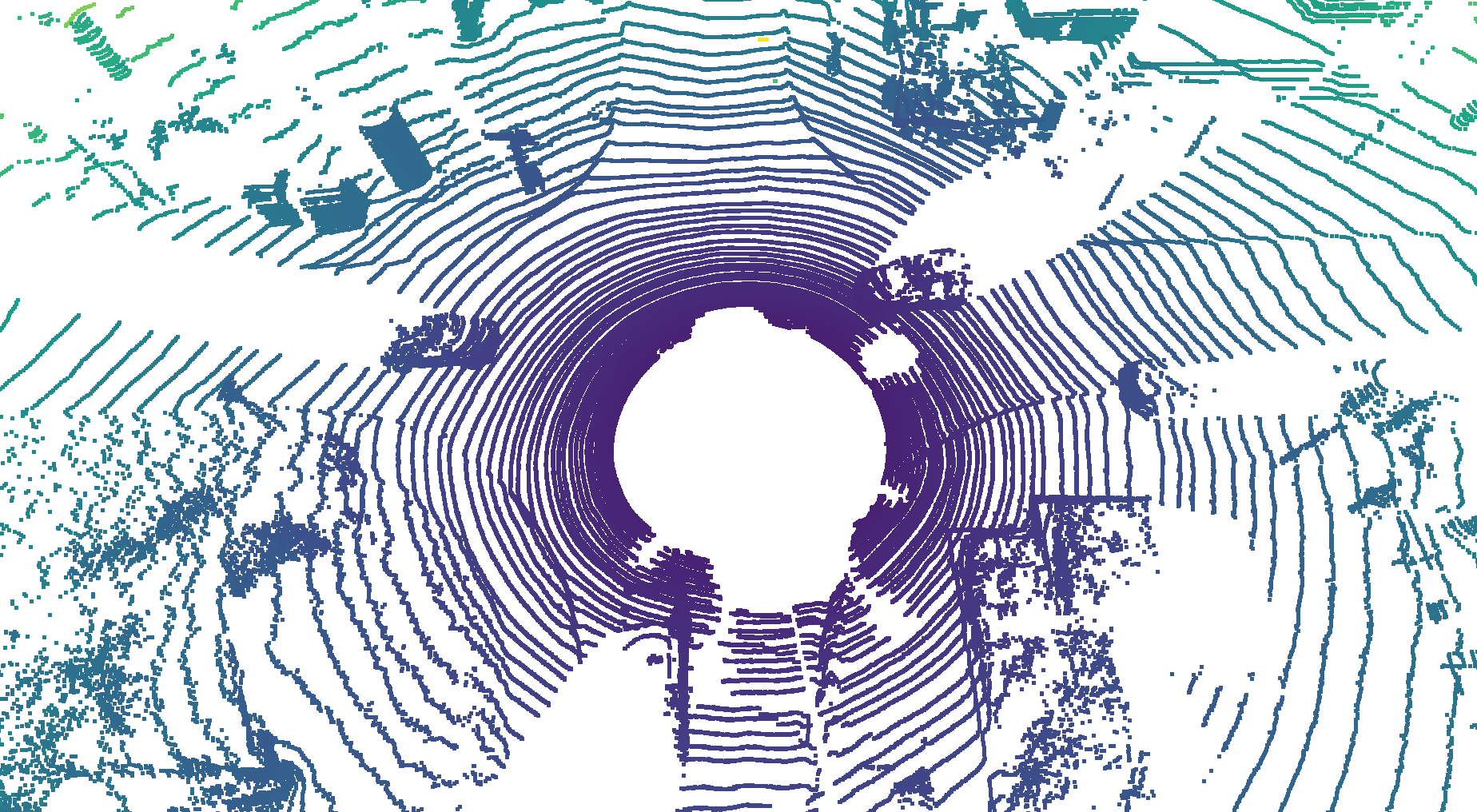}}
	\subfigure[Ground Truth Labels]{\includegraphics[width=0.42\linewidth]{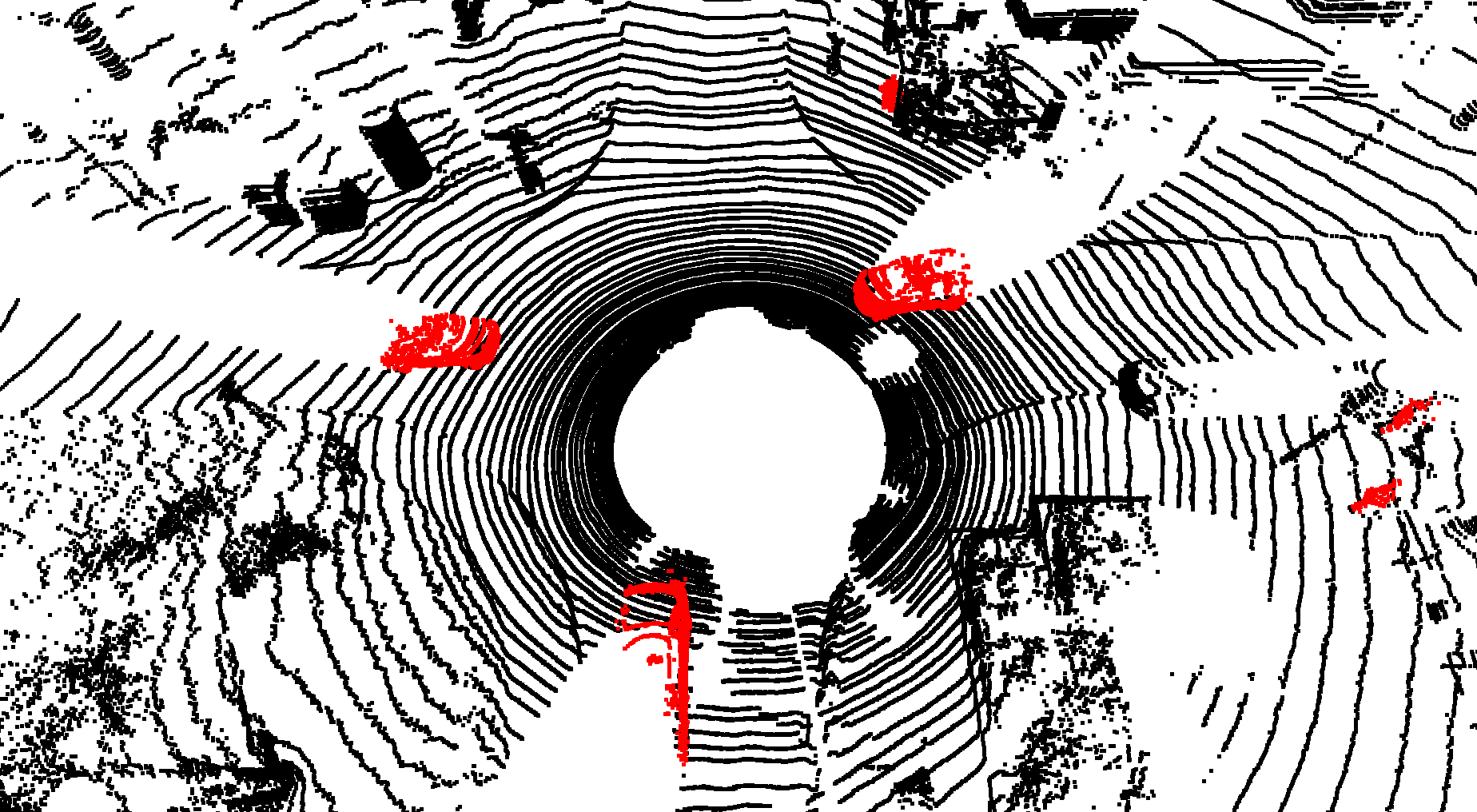}}
	\subfigure[SalsaNext (retrained)]{\includegraphics[width=0.42\linewidth]{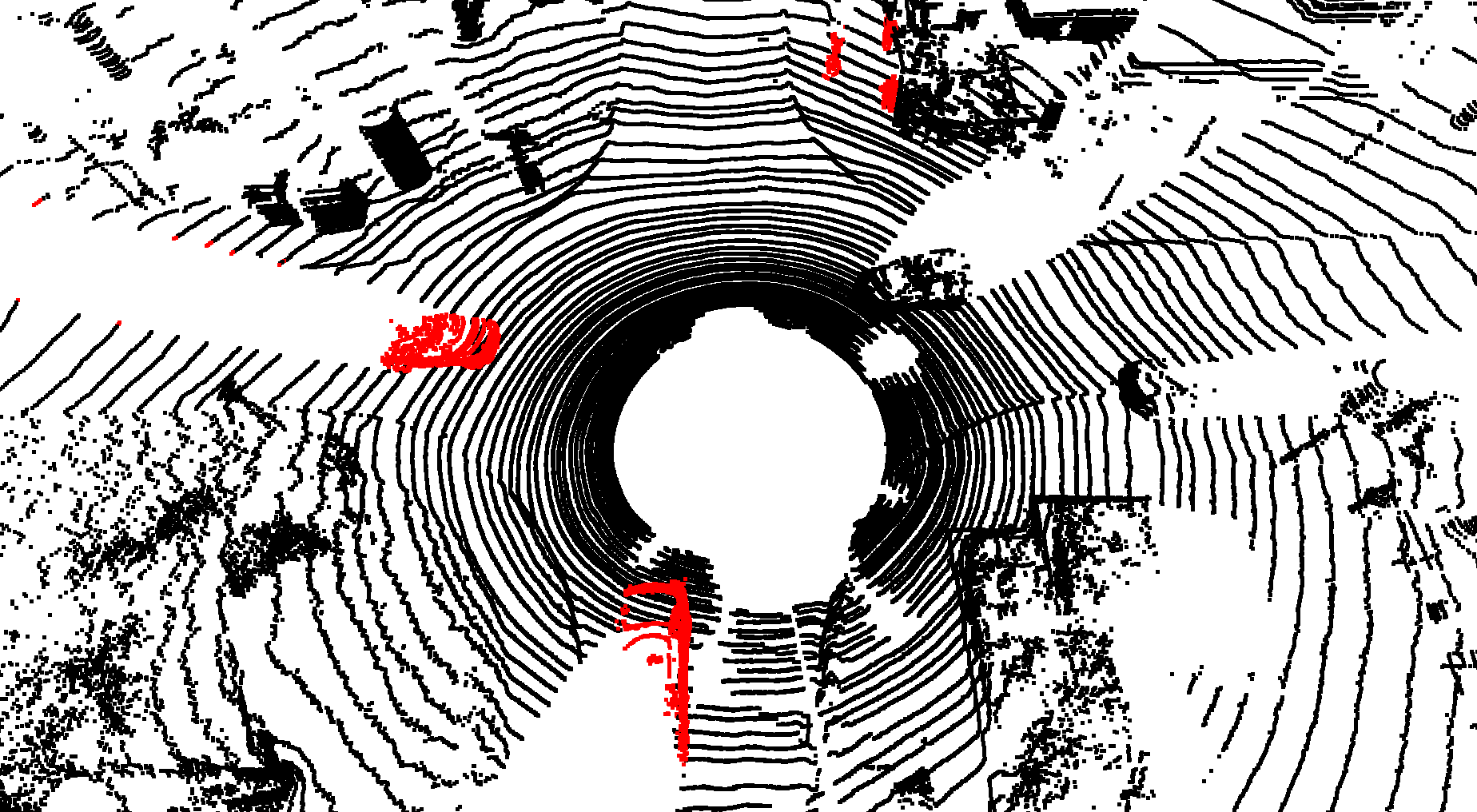}}
	\subfigure[SalsaNext+ N = 8 + Semantics (Ours)]{\includegraphics[width=0.42\linewidth]{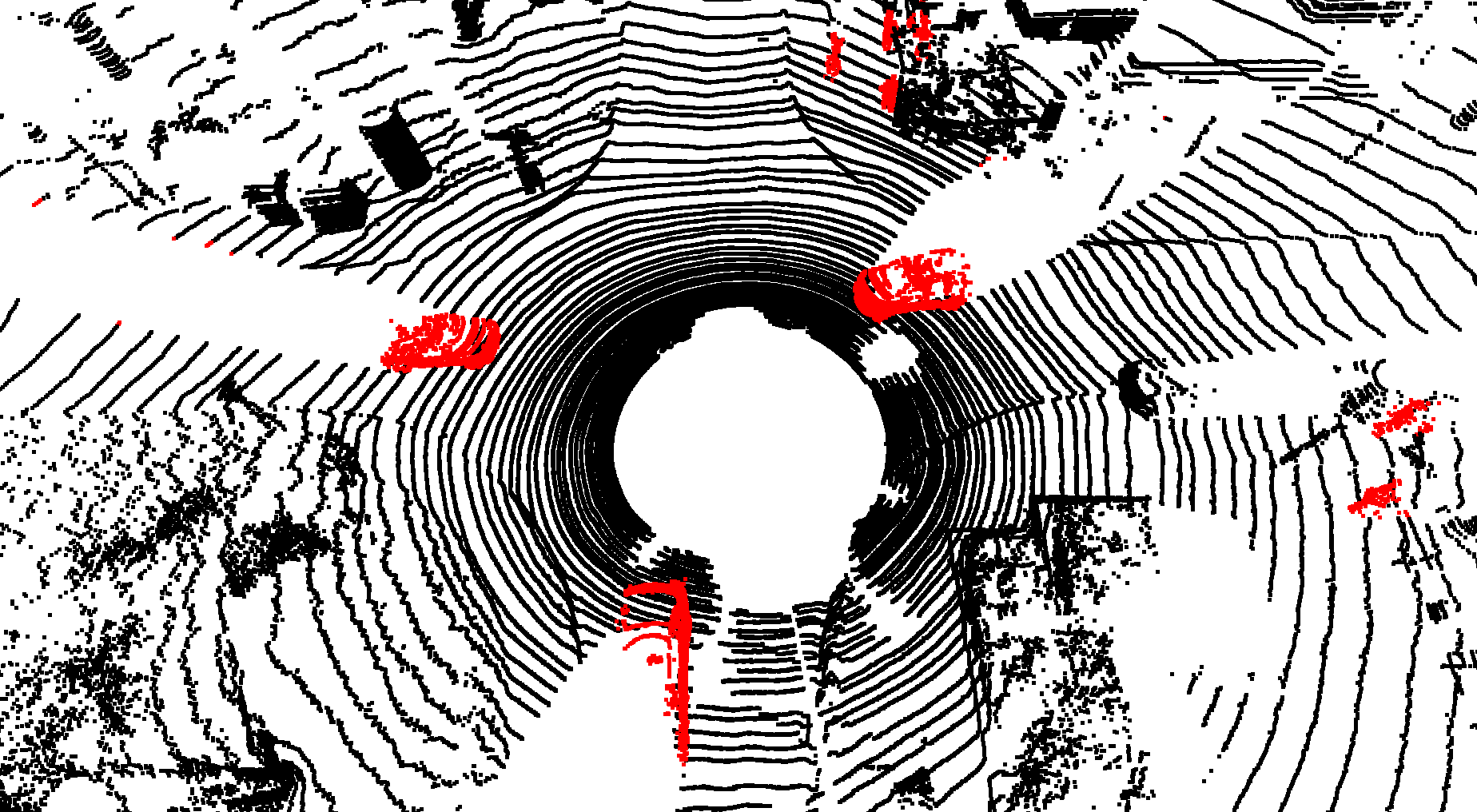}}
	\caption{Qualitative results shown as point clouds. (a) shows the raw point cloud with points colored depending on the range from purple (near) to yellow (far).  (b) shows the ground truth, and (c,d) prediction results, where red points correspond to the class moving.}
	\label{fig:points}
\end{figure*}

\figref{fig:qualitative} and \figref{fig:points} show the qualitative results on range images and LiDAR scans respectively in a very challenging situation, where the car is at the intersection and there are both a lot of moving objects and static objects.
Our method can distinguish moving and static points even when some of the moving objects are moving slowly and other methods fail to detect this.

\subsection{Applications}

Two obvious applications of our proposed method are LiDAR-based odometry/SLAM as well as 3D mapping.
Here, we show the effectiveness of our method by using the MOS predictions as masks for the input LiDAR scans, which removes effectively all points belonging to moving objects. No further tweaks have been employed.
We use our best setup for the MOS, \ie, our approach extending SalsaNext with $N=8$ residual images and semantics.

\subsubsection{Odometry/SLAM}

For the LiDAR-based odometry experiments, we use an off-the-shelf SLAM approach~\cite{behley2018rss} and apply our MOS method before feeding the point cloud into the SLAM pipeline.
We compare the improved odometry results to both the original approach, called SuMa, and our semantic-enhanced approach, SuMa++~\cite{chen2019iros}.
We evaluate these odometry methods, SuMa, SuMa++, and SuMa+MOS on the KITTI odometry benchmark~\cite{geiger2012cvpr}.

The quantitative results are shown in~\tabref{tab:odometry}.
We can see that, by simply applying our MOS predictions as a pre-processing mask, the odometry results are improved in both the KITTI training and test data and even slightly better than the well-designed semantic-enhanced SuMa.

\begin{table}[t]
	\centering
	\caption{KITTI Odometry Benchmark Results}
	\begin{tabular}{cC{1.5cm}C{1.5cm}C{1.5cm}}
		\toprule
		\multirow{3}{*}{Split} & \multicolumn{3}{c}{Approach}                                                                                                                                                                                                                                                                        \\
		\cmidrule{2-4}
		                       & SuMa                         & SuMa++        & SuMa+MOS                                                                                                                                                                                                                                             \\
		\midrule
		Train (Seq. 00-10)      & $0.36$/$0.83$                & $0.29$/$0.70$ & $\mathbf{0.29}$/$\mathbf{0.66}$                                                                                                                                                                                                                      \\
		Test (Seq. 11-21)      & $0.34$/$1.39$                & $0.34$/$1.06$ & $\mathbf{0.33}$/$\mathbf{0.99}$                                                                                                                                                                                                                      \\
		\bottomrule
		\multicolumn{4}{p{0.9\linewidth}}{\vspace{0.01cm}\scriptsize{Relative errors averaged over trajectories of $100$ to $800$\,m length: relative rotational error in $\mathrm{degrees}$ per $100$\,m / relative translational error in $\%$. 
}} \\
	\end{tabular}
	\label{tab:odometry}
	\vspace{-0.3cm}
\end{table}

\subsubsection{3D Mapping}
\begin{figure*}[t]
    \vspace{0.2cm}
	\centering
	\subfigure[Raw point clouds]{\includegraphics[width=0.45\linewidth]{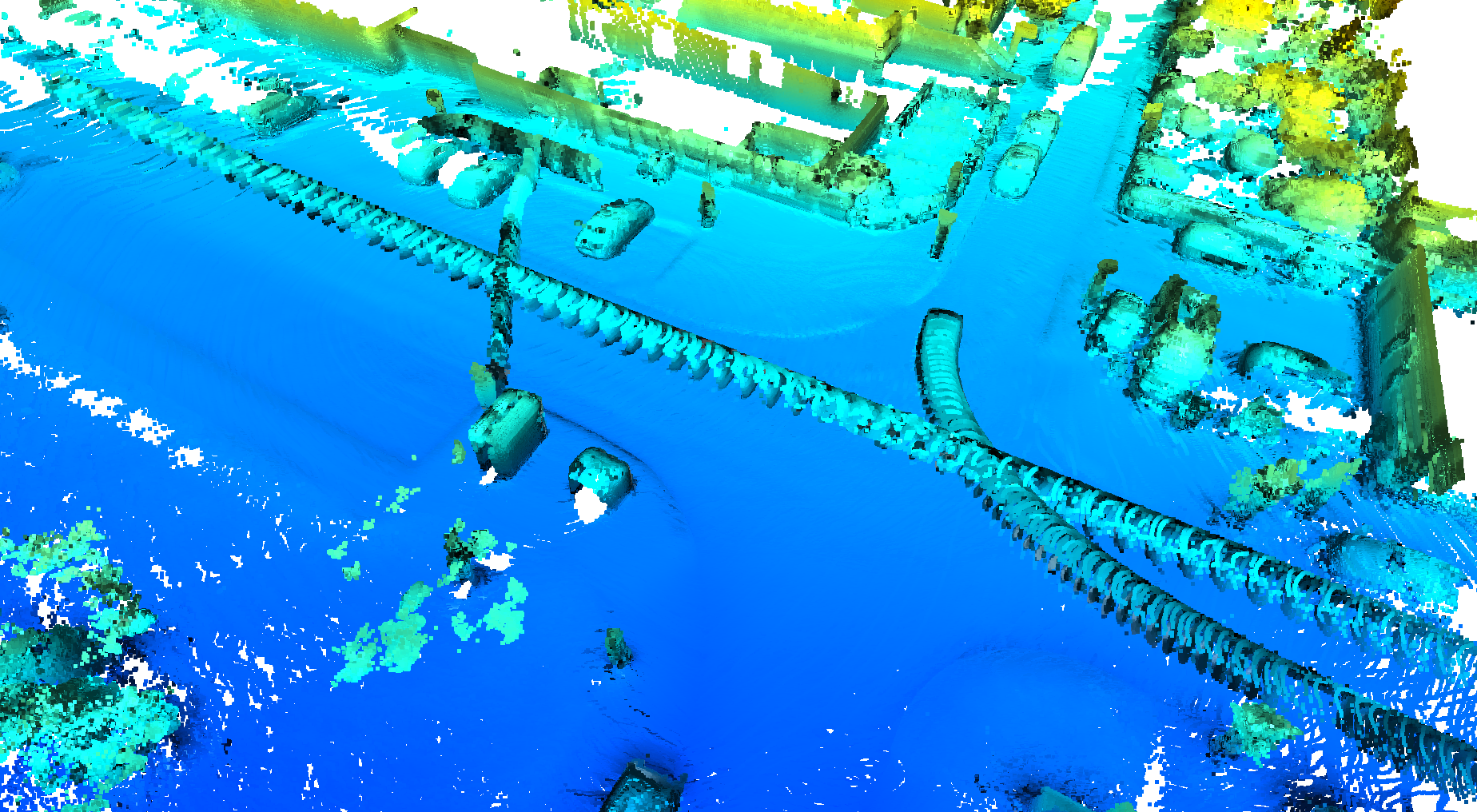}}\hspace{0.75cm} 
	\subfigure[Point clouds with moving segments removed]{\includegraphics[width=0.45\linewidth]{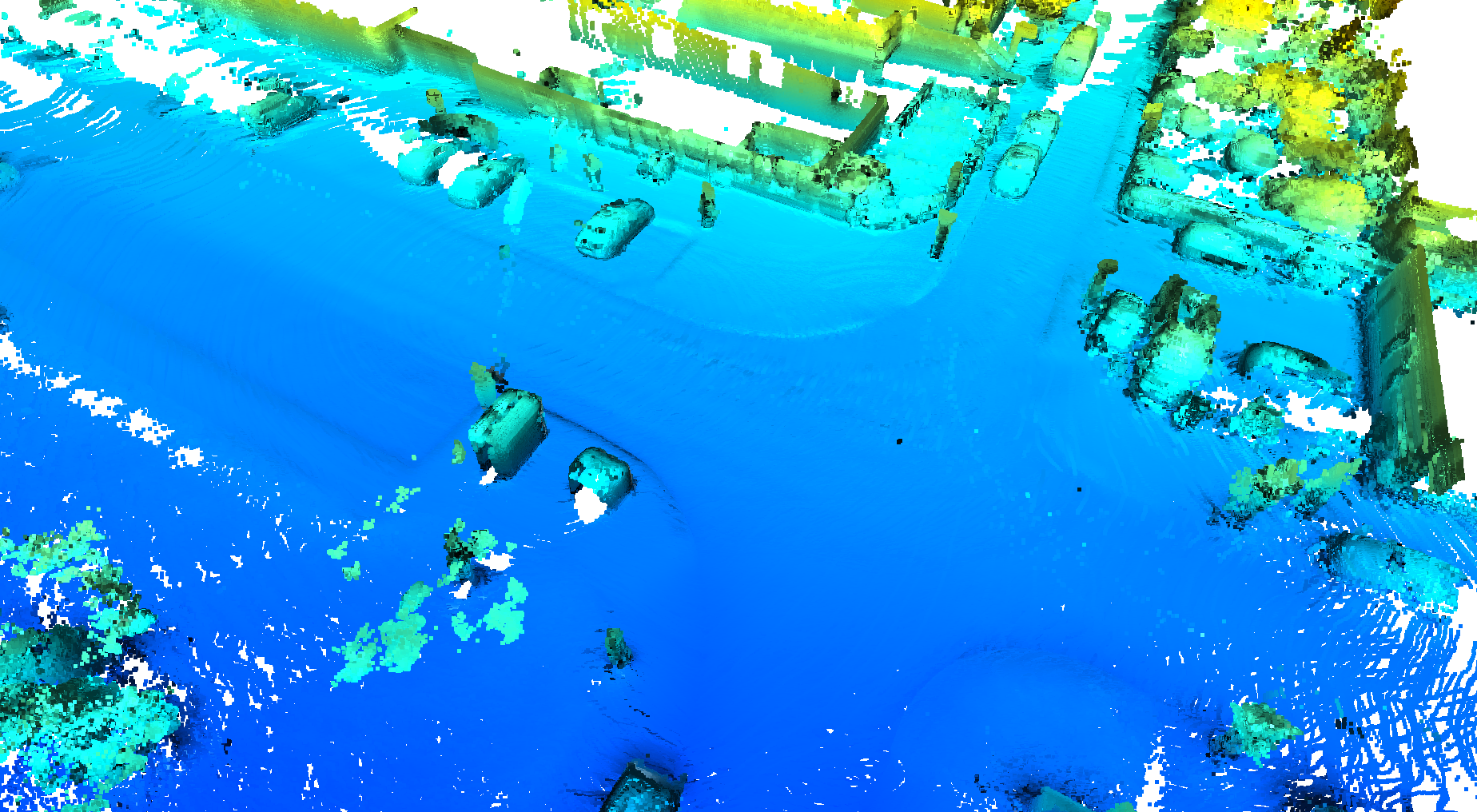}}
	\caption{Mapping Results on Sequence 08, Frame 3960-4070, where we show the accumulated point cloud (a) without removing segments and (b) when we remove the segments predicted as moving.}
	\label{fig:map}
\end{figure*}

As shown in~\figref{fig:map}, we compare the aggregated point cloud maps (a) directly with the raw LiDAR scans, (b) with the cleaned LiDAR scans by applying our MOS predictions as masks.
We use the Open3D library~\cite{zhou2018arxiv} to visualize the mapping results.
As can be seen, there are moving objects present that pollute the map, which might have adversarial effects, when used for localization or path planning.
By using our MOS predictions as masks, we can effectively remove these artifacts and get a clean map.
Note that, here we show two direct use cases of our MOS approach without any further optimizations employed. 

\subsection{Runtime}
\label{sec:runtime}
The runtime is evaluated on sequence 08 with an Intel i7-8700 with 3.2 GHz and a single Nvidia Quadro P6000 graphic card.
It takes around 10\,ms on average to estimate the odometry and generate the residual image.
Since we only change the input of each network while keeping the architecture the same, the inference time is nearly the same as before, specifically 75\,ms for RangeNet++, 42\,ms for SalsaNext, and 21\,ms for MINet.
In case of using semantics for the MOS, we can run a second full semantics network in parallel.

As the odometry history for the SLAM is available, we need to estimate the pose and generate the residual images only once for every incoming frame.
In sum, using our method for LiDAR-based odometry takes approx. 51\,ms per scan (=20\,Hz) using SalsaNext, which is faster than the frame rate of a typical LiDAR sensor, \ie, 10\,Hz.

\section{Conclusion}
\label{sec:conclusion}

In this paper, we presented a novel and effective approach to achieve LiDAR-based moving object segmentation in an end-to-end online fashion.
Our method exploits neural networks and sequential information, which allows our approach to successfully distinguish between moving and static objects.
Our method is based on range projections and thus fast and can directly improve existing SLAM and mapping systems.
The experiments suggest that our method achieves good performance on MOS and outperformed several state-of-the-art approaches.
We also propose a new benchmark for LiDAR-based MOS and used it to evaluate our approach, also allowing further comparisons with future MOS systems.

\section*{Acknowledgments}
Thanks to Thomas~L\"abe, Ignacio~Vizzo, Nived~Chebrolu, Tiziano~Guadagnino, and Haofei~Kuang for fruitful discussions and early support.

\bibliographystyle{plain_abbrv}

\bibliography{glorified,new}

\end{document}

%% file: stachnisslab-latex.tex
\usepackage{graphics}           
\usepackage{times}              
\usepackage{amsmath}            
\usepackage{amssymb}            
\usepackage{graphicx}
\usepackage{algorithm}
\usepackage[noend]{algpseudocode}
\usepackage{booktabs}
\usepackage{color}
\definecolor{instructioncolor}{rgb}{.5,.5,.5}

\usepackage[font=small]{caption}

\def\secref#1{Sec.~\ref{#1}}
\def\figref#1{Fig.~\ref{#1}}
\def\tabref#1{Tab.~\ref{#1}}
\def\eqref#1{Eq.~(\ref{#1})}

\makeatletter
\usepackage{xspace}
\DeclareRobustCommand\onedot{\futurelet\@let@token\@onedot}
\def\@onedot{\ifx\@let@token.\else.\null\fi\xspace}
\def\eg{e.g\onedot} 
\def\ie{i.e\onedot} 
 
 \def\vs{vs\onedot}
 
\def\etal{{et al}\onedot}
\makeatother

\usepackage{array}
\newcolumntype{L}[1]{>{\raggedright\let\newline\\\arraybackslash\hspace{0pt}}m{#1}}
\newcolumntype{C}[1]{>{\centering\let\newline\\\arraybackslash\hspace{0pt}}m{#1}}
\newcolumntype{R}[1]{>{\raggedleft\let\newline\\\arraybackslash\hspace{0pt}}m{#1}}

%% file: stachnisslab-math.tex
\newcommand{\RR}{\mathbb{R}}

\newcommand{\abs}[1]{|#1|}

\renewcommand{\b}[1]{\mbox{\boldmath$#1$}}

\renewcommand{\v}[1]{{\b #1}} 

\newcommand{\m}[1]{{\mbox{{\sffamily\slshape{#1\/}}}}}



%% file: pics/motivation_smaller.pdf_tex
\begingroup%
  \makeatletter%
  \providecommand\color[2][]{%
    \errmessage{(Inkscape) Color is used for the text in Inkscape, but the package 'color.sty' is not loaded}%
    \renewcommand\color[2][]{}%
  }%
  \providecommand\transparent[1]{%
    \errmessage{(Inkscape) Transparency is used (non-zero) for the text in Inkscape, but the package 'transparent.sty' is not loaded}%
    \renewcommand\transparent[1]{}%
  }%
  \providecommand\rotatebox[2]{#2}%
  \newcommand*\fsize{\dimexpr\f@size pt\relax}%
  \newcommand*\lineheight[1]{\fontsize{\fsize}{#1\fsize}\selectfont}%
  \ifx\svgwidth\undefined%
    \setlength{\unitlength}{264.21629333bp}%
    \ifx\svgscale\undefined%
      \relax%
    \else%
      \setlength{\unitlength}{\unitlength * \real{\svgscale}}%
    \fi%
  \else%
    \setlength{\unitlength}{\svgwidth}%
  \fi%
  \global\let\svgwidth\undefined%
  \global\let\svgscale\undefined%
  \makeatother%
  \begin{picture}(1,0.70012748)%
    \lineheight{1}%
    \setlength\tabcolsep{0pt}%
    \put(0,0){\includegraphics[width=\unitlength,page=1]{motivation_smaller.pdf}}%
    \put(0.12791474,0.41460534){\color[rgb]{0,0,0}\makebox(0,0)[lt]{\lineheight{1.25}\smash{\begin{tabular}[t]{l}Raw Point Cloud\\\end{tabular}}}}%
    \put(0.54165141,0.41460534){\color[rgb]{0,0,0}\makebox(0,0)[lt]{\lineheight{1.25}\smash{\begin{tabular}[t]{l}Segmented Point Cloud\\\end{tabular}}}}%
    \put(0.36379817,0.25233206){\color[rgb]{0,0,0}\makebox(0,0)[lt]{\lineheight{1.25}\smash{\begin{tabular}[t]{l}Range Images\\\end{tabular}}}}%
    \put(0.3568865,0.12444629){\color[rgb]{0,0,0}\makebox(0,0)[lt]{\lineheight{1.25}\smash{\begin{tabular}[t]{l}Our Predictions\\\end{tabular}}}}%
    \put(0.32011785,0.00042875){\color[rgb]{0,0,0}\makebox(0,0)[lt]{\lineheight{1.25}\smash{\begin{tabular}[t]{l}Ground Truth Labels\\\end{tabular}}}}%
    \put(0,0){\includegraphics[width=\unitlength,page=2]{motivation_smaller.pdf}}%
  \end{picture}%
\endgroup%

%% file: pics/overview.pdf_tex
\begingroup%
  \makeatletter%
  \providecommand\color[2][]{%
    \errmessage{(Inkscape) Color is used for the text in Inkscape, but the package 'color.sty' is not loaded}%
    \renewcommand\color[2][]{}%
  }%
  \providecommand\transparent[1]{%
    \errmessage{(Inkscape) Transparency is used (non-zero) for the text in Inkscape, but the package 'transparent.sty' is not loaded}%
    \renewcommand\transparent[1]{}%
  }%
  \providecommand\rotatebox[2]{#2}%
  \newcommand*\fsize{\dimexpr\f@size pt\relax}%
  \newcommand*\lineheight[1]{\fontsize{\fsize}{#1\fsize}\selectfont}%
  \ifx\svgwidth\undefined%
    \setlength{\unitlength}{429.33519258bp}%
    \ifx\svgscale\undefined%
      \relax%
    \else%
      \setlength{\unitlength}{\unitlength * \real{\svgscale}}%
    \fi%
  \else%
    \setlength{\unitlength}{\svgwidth}%
  \fi%
  \global\let\svgwidth\undefined%
  \global\let\svgscale\undefined%
  \makeatother%
  \begin{picture}(1,0.3949719)%
    \lineheight{1}%
    \setlength\tabcolsep{0pt}%
    \put(0,0){\includegraphics[width=\unitlength,page=1]{overview.pdf}}%
    \put(0.28084057,0.07034673){\color[rgb]{0,0,0}\makebox(0,0)[lt]{\lineheight{1.25}\smash{\begin{tabular}[t]{l}$r,x,y,z,e$\end{tabular}}}}%
    \put(0.37803677,0.12034574){\color[rgb]{0,0,0}\makebox(0,0)[lt]{\lineheight{1.25}\smash{\begin{tabular}[t]{l}$r$\end{tabular}}}}%
    \put(0.48583852,0.31180446){\color[rgb]{0,0,0}\makebox(0,0)[lt]{\lineheight{1.25}\smash{\begin{tabular}[t]{l}$d_{2}^{0}$\end{tabular}}}}%
    \put(0.48567809,0.1990818){\color[rgb]{0,0,0}\makebox(0,0)[lt]{\lineheight{1.25}\smash{\begin{tabular}[t]{l}$d_{1}^{0}$\end{tabular}}}}%
    \put(0.94073329,0.06495988){\color[rgb]{0,0,0}\makebox(0,0)[t]{\lineheight{1.25}\smash{\begin{tabular}[t]{c}Binary\\Output Mask\end{tabular}}}}%
    \put(0.72962411,0.13462814){\color[rgb]{0,0,0}\makebox(0,0)[t]{\lineheight{1.25}\smash{\begin{tabular}[t]{c}CNN\end{tabular}}}}%
    \put(0.07737141,0.24952164){\color[rgb]{0,0,0}\makebox(0,0)[lt]{\lineheight{1.25}\smash{\begin{tabular}[t]{l}$\m{S}_{2}$\end{tabular}}}}%
    \put(0.07737141,0.13500543){\color[rgb]{0,0,0}\makebox(0,0)[lt]{\lineheight{1.25}\smash{\begin{tabular}[t]{l}$\m{S}_{1}$\end{tabular}}}}%
    \put(0.07737141,0.02408047){\color[rgb]{0,0,0}\makebox(0,0)[lt]{\lineheight{1.25}\smash{\begin{tabular}[t]{l}$\m{S}_0$\end{tabular}}}}%
    \put(0,0){\includegraphics[width=\unitlength,page=2]{overview.pdf}}%
    \put(0.76593584,0.20869507){\color[rgb]{0,0,0}\makebox(0,0)[lt]{\lineheight{1.25}\smash{\begin{tabular}[t]{l}Residual Images\end{tabular}}}}%
    \put(0,0){\includegraphics[width=\unitlength,page=3]{overview.pdf}}%
    \put(0.24149909,0.31926632){\color[rgb]{0,0,0}\makebox(0,0)[lt]{\lineheight{1.25}\smash{\begin{tabular}[t]{l}$\Pi$\end{tabular}}}}%
    \put(0.24149909,0.20193746){\color[rgb]{0,0,0}\makebox(0,0)[lt]{\lineheight{1.25}\smash{\begin{tabular}[t]{l}$\Pi$\end{tabular}}}}%
    \put(0.24149909,0.0897341){\color[rgb]{0,0,0}\makebox(0,0)[lt]{\lineheight{1.25}\smash{\begin{tabular}[t]{l}$\Pi$\end{tabular}}}}%
    \put(0,0){\includegraphics[width=\unitlength,page=4]{overview.pdf}}%
    \put(0.19552368,0.191141){\color[rgb]{0,0,0}\makebox(0,0)[lt]{\lineheight{1.25}\smash{\begin{tabular}[t]{l}$\v{T}_{1}^{0}$\end{tabular}}}}%
    \put(0.19622421,0.30727165){\color[rgb]{0,0,0}\makebox(0,0)[lt]{\lineheight{1.25}\smash{\begin{tabular}[t]{l}$\v{T}_{2}^{0}$\end{tabular}}}}%
    \put(0,0){\includegraphics[width=\unitlength,page=5]{overview.pdf}}%
    \put(0.41345736,0.18443055){\color[rgb]{0,0,0}\makebox(0,0)[lt]{\lineheight{1.25}\smash{\begin{tabular}[t]{l}$|\Delta r|$\end{tabular}}}}%
    \put(0.41301348,0.29643912){\color[rgb]{0,0,0}\makebox(0,0)[lt]{\lineheight{1.25}\smash{\begin{tabular}[t]{l}$|\Delta r|$\end{tabular}}}}%
    \put(0,0){\includegraphics[width=\unitlength,page=6]{overview.pdf}}%
  \end{picture}%
\endgroup%

%% file: pics/residual_images.pdf_tex
\begingroup%
  \makeatletter%
  \providecommand\color[2][]{%
    \errmessage{(Inkscape) Color is used for the text in Inkscape, but the package 'color.sty' is not loaded}%
    \renewcommand\color[2][]{}%
  }%
  \providecommand\transparent[1]{%
    \errmessage{(Inkscape) Transparency is used (non-zero) for the text in Inkscape, but the package 'transparent.sty' is not loaded}%
    \renewcommand\transparent[1]{}%
  }%
  \providecommand\rotatebox[2]{#2}%
  \newcommand*\fsize{\dimexpr\f@size pt\relax}%
  \newcommand*\lineheight[1]{\fontsize{\fsize}{#1\fsize}\selectfont}%
  \ifx\svgwidth\undefined%
    \setlength{\unitlength}{331.39421082bp}%
    \ifx\svgscale\undefined%
      \relax%
    \else%
      \setlength{\unitlength}{\unitlength * \real{\svgscale}}%
    \fi%
  \else%
    \setlength{\unitlength}{\svgwidth}%
  \fi%
  \global\let\svgwidth\undefined%
  \global\let\svgscale\undefined%
  \makeatother%
  \begin{picture}(1,0.57131766)%
    \lineheight{1}%
    \setlength\tabcolsep{0pt}%
    \put(0,0){\includegraphics[width=\unitlength,page=1]{residual_images.pdf}}%
    \put(0.90520611,0.52943492){\makebox(0,0)[lt]{\lineheight{1.25}\smash{\begin{tabular}[t]{l}Current\end{tabular}}}}%
    \put(0,0){\includegraphics[width=\unitlength,page=2]{residual_images.pdf}}%
    \put(0.91361044,0.416234){\makebox(0,0)[lt]{\lineheight{1.25}\smash{\begin{tabular}[t]{l}$j = 1$\end{tabular}}}}%
    \put(0,0){\includegraphics[width=\unitlength,page=3]{residual_images.pdf}}%
    \put(0.91361044,0.35943365){\makebox(0,0)[lt]{\lineheight{1.25}\smash{\begin{tabular}[t]{l}$j = 2$\end{tabular}}}}%
    \put(0,0){\includegraphics[width=\unitlength,page=4]{residual_images.pdf}}%
    \put(0.91361044,0.30295157){\makebox(0,0)[lt]{\lineheight{1.25}\smash{\begin{tabular}[t]{l}$j = 3$\end{tabular}}}}%
    \put(0,0){\includegraphics[width=\unitlength,page=5]{residual_images.pdf}}%
    \put(0.91361044,0.24678772){\makebox(0,0)[lt]{\lineheight{1.25}\smash{\begin{tabular}[t]{l}$j = 4$\end{tabular}}}}%
    \put(0,0){\includegraphics[width=\unitlength,page=6]{residual_images.pdf}}%
    \put(0.91361044,0.1903056){\makebox(0,0)[lt]{\lineheight{1.25}\smash{\begin{tabular}[t]{l}$j = 5$\end{tabular}}}}%
    \put(0,0){\includegraphics[width=\unitlength,page=7]{residual_images.pdf}}%
    \put(0.91361044,0.13350529){\makebox(0,0)[lt]{\lineheight{1.25}\smash{\begin{tabular}[t]{l}$j = 6$\end{tabular}}}}%
    \put(0,0){\includegraphics[width=\unitlength,page=8]{residual_images.pdf}}%
    \put(0.91361044,0.07734143){\makebox(0,0)[lt]{\lineheight{1.25}\smash{\begin{tabular}[t]{l}$j = 7$\end{tabular}}}}%
    \put(0,0){\includegraphics[width=\unitlength,page=9]{residual_images.pdf}}%
    \put(0.91361044,0.02054114){\makebox(0,0)[lt]{\lineheight{1.25}\smash{\begin{tabular}[t]{l}$j = 8$\end{tabular}}}}%
    \put(0,0){\includegraphics[width=\unitlength,page=10]{residual_images.pdf}}%
    \put(0.91155358,0.47185662){\makebox(0,0)[lt]{\lineheight{1.25}\smash{\begin{tabular}[t]{l}Labels\end{tabular}}}}%
  \end{picture}%
\endgroup%

%% file: pics/qualitative.pdf_tex
\begingroup%
  \makeatletter%
  \providecommand\color[2][]{%
    \errmessage{(Inkscape) Color is used for the text in Inkscape, but the package 'color.sty' is not loaded}%
    \renewcommand\color[2][]{}%
  }%
  \providecommand\transparent[1]{%
    \errmessage{(Inkscape) Transparency is used (non-zero) for the text in Inkscape, but the package 'transparent.sty' is not loaded}%
    \renewcommand\transparent[1]{}%
  }%
  \providecommand\rotatebox[2]{#2}%
  \newcommand*\fsize{\dimexpr\f@size pt\relax}%
  \newcommand*\lineheight[1]{\fontsize{\fsize}{#1\fsize}\selectfont}%
  \ifx\svgwidth\undefined%
    \setlength{\unitlength}{2029.20410156bp}%
    \ifx\svgscale\undefined%
      \relax%
    \else%
      \setlength{\unitlength}{\unitlength * \real{\svgscale}}%
    \fi%
  \else%
    \setlength{\unitlength}{\svgwidth}%
  \fi%
  \global\let\svgwidth\undefined%
  \global\let\svgscale\undefined%
  \makeatother%
  \begin{picture}(1,0.27543575)%
    \lineheight{1}%
    \setlength\tabcolsep{0pt}%
    \put(0,0){\includegraphics[width=\unitlength,page=1]{qualitative.pdf}}%
    \put(0.19518185,0.21459959){\makebox(0,0)[lt]{\lineheight{1.25}\smash{\begin{tabular}[t]{l}Range Image\end{tabular}}}}%
    \put(0.13402771,-0.00145159){\makebox(0,0)[lt]{\lineheight{1.25}\smash{\begin{tabular}[t]{l}SalsaNext (moveable classes)\end{tabular}}}}%
    \put(0.15548923,0.07113038){\makebox(0,0)[lt]{\lineheight{1.25}\smash{\begin{tabular}[t]{l}SceneFlow + Semantics\end{tabular}}}}%
    \put(0.1647654,0.14384375){\makebox(0,0)[lt]{\lineheight{1.25}\smash{\begin{tabular}[t]{l}Diff+RG + Semantics\end{tabular}}}}%
    \put(0.68456351,0.21414481){\makebox(0,0)[lt]{\lineheight{1.25}\smash{\begin{tabular}[t]{l}Ground Truth Labels\end{tabular}}}}%
    \put(0.68031885,0.1426079){\makebox(0,0)[lt]{\lineheight{1.25}\smash{\begin{tabular}[t]{l}SalsaNext (retrained)\end{tabular}}}}%
    \put(0.66477675,0.07113038){\makebox(0,0)[lt]{\lineheight{1.25}\smash{\begin{tabular}[t]{l}SalsaNext + N = 1 (Ours)\end{tabular}}}}%
    \put(0.61536039,0.00020935){\makebox(0,0)[lt]{\lineheight{1.25}\smash{\begin{tabular}[t]{l}SalsaNext + N = 8 + Semantics (Ours)\end{tabular}}}}%
  \end{picture}%
\endgroup%